\documentclass[letterpaper, 10 pt, conference]{ieeeconf}  
\IEEEoverridecommandlockouts                              

\usepackage{graphics} 
\usepackage{epsfig} 
\usepackage{mathptmx} 
\usepackage{times} 
\usepackage{amsmath} 
\usepackage{amssymb}  

\usepackage[utf8]{inputenc} 
\usepackage[T1]{fontenc}    
\usepackage{hyperref}       
\usepackage{url}            
\usepackage{booktabs}       
\usepackage{amsfonts}       
\usepackage{nicefrac}       
\usepackage{microtype}      
\usepackage{lipsum}
\usepackage{tikz}
\usepackage{multirow}
\usepackage{amsmath}
\usepackage{graphics}
\usepackage{subcaption}
\usepackage{makecell}
\usepackage{hyperref}
\usepackage{xcolor}
\usepackage{rotating}
\usepackage{pgfplots}
\pgfplotsset{compat=1.8}
\usepackage{eurosym}      
\usepackage{filecontents} 
\usepackage{tabu}
\usepackage{tikz}
\usepackage{rotating}
\usepackage{graphicx}
\usepackage{gensymb}
\usepackage{todonotes}
\def\checkmark{\tikz\fill[scale=0.4](0,.35) -- (.25,0) -- (1,.7) -- (.25,.15) -- cycle;}

\title{\LARGE \bf
S3Net: 3D LiDAR Sparse Semantic Segmentation Network
}

\author{Ran Cheng*, Ryan Razani*, Yuan Ren and Liu Bingbing\\ 
Huawei Noah's Ark Lab, Toronto, Canada\\
\texttt{\{ran.cheng, ryan.razani, yuan.ren3, liu.bingbing\}@huawei.com}
\thanks{$^*$ indicates equal contribution.}
}

\begin{document}

\maketitle
\thispagestyle{empty}
\pagestyle{empty}


\begin{abstract}

Semantic Segmentation is a crucial component in the
perception systems of many applications, such as robotics and  autonomous driving that rely on accurate environmental perception and understanding. In literature, several approaches are introduced to attempt LiDAR semantic segmentation task, such as projection-based (range-view or birds-eye-view), and voxel-based approaches. However, they either abandon the valuable 3D topology and geometric relations and suffer from information loss introduced in the projection process or are inefficient. Therefore, there is a need for accurate models capable of processing the 3D driving-scene point cloud in 3D space. In this paper, we propose S3Net, a novel convolutional neural network for LiDAR point cloud semantic segmentation. It adopts an encoder-decoder backbone that consists of Sparse Intra-channel Attention Module (SIntraAM), and Sparse Inter-channel Attention Module (SInterAM) to emphasize the fine details of both within each feature map and among nearby feature maps. To extract the global contexts in deeper layers, we introduce Sparse Residual Tower based upon sparse convolution that suits varying sparsity of LiDAR point cloud. In addition, geo-aware anisotropic loss is leveraged to emphasize the semantic boundaries and penalize the noise within each predicted regions, leading to a robust prediction. Our experimental results show that the proposed method leads to a large improvement (12\%) compared to its baseline counterpart (MinkNet42 \cite{choy20194d}) on SemanticKITTI  \cite{DBLP:conf/iccv/BehleyGMQBSG19} test set and achieves state-of-the-art mIoU accuracy of semantic segmentation approaches.

\end{abstract}

\section{INTRODUCTION}
Today, the increasing demand for an accurate perception system which finds different applications such as robotics \cite{thrun2006stanley} and autonomous vehicles \cite{li2016vehicle}, motivates the research of improved $3$D deep learning networks that can perceive the semantics of outdoor scene. 
Various sensors can be leveraged for scene understanding task, however LiDAR sensor has become the main sensing modalities in perception tasks such as object detection and semantic segmentation.
$3$D LiDAR sensor provides a rich $3$D geometry representation that mimics the real-world.
Semantic segmentation is a fundamental task of scene understanding, which helps gaining a rich understanding of the scene. The purpose of semantic segmentation of LiDAR point clouds is 
to predict the class label of each $3$D point in a given LiDAR scan. In the context of autonomous driving, however, object detection or semantic segmentation is not totally independent. As class label for object of interest can be generated by semantic segmentation. Thus, segmentation can act as an intermediate step to enhance other downstream perception tasks, like object detection and tracking. 

\begin{figure}[htbp!]
    \centering
    \includegraphics[width=0.5\textwidth]{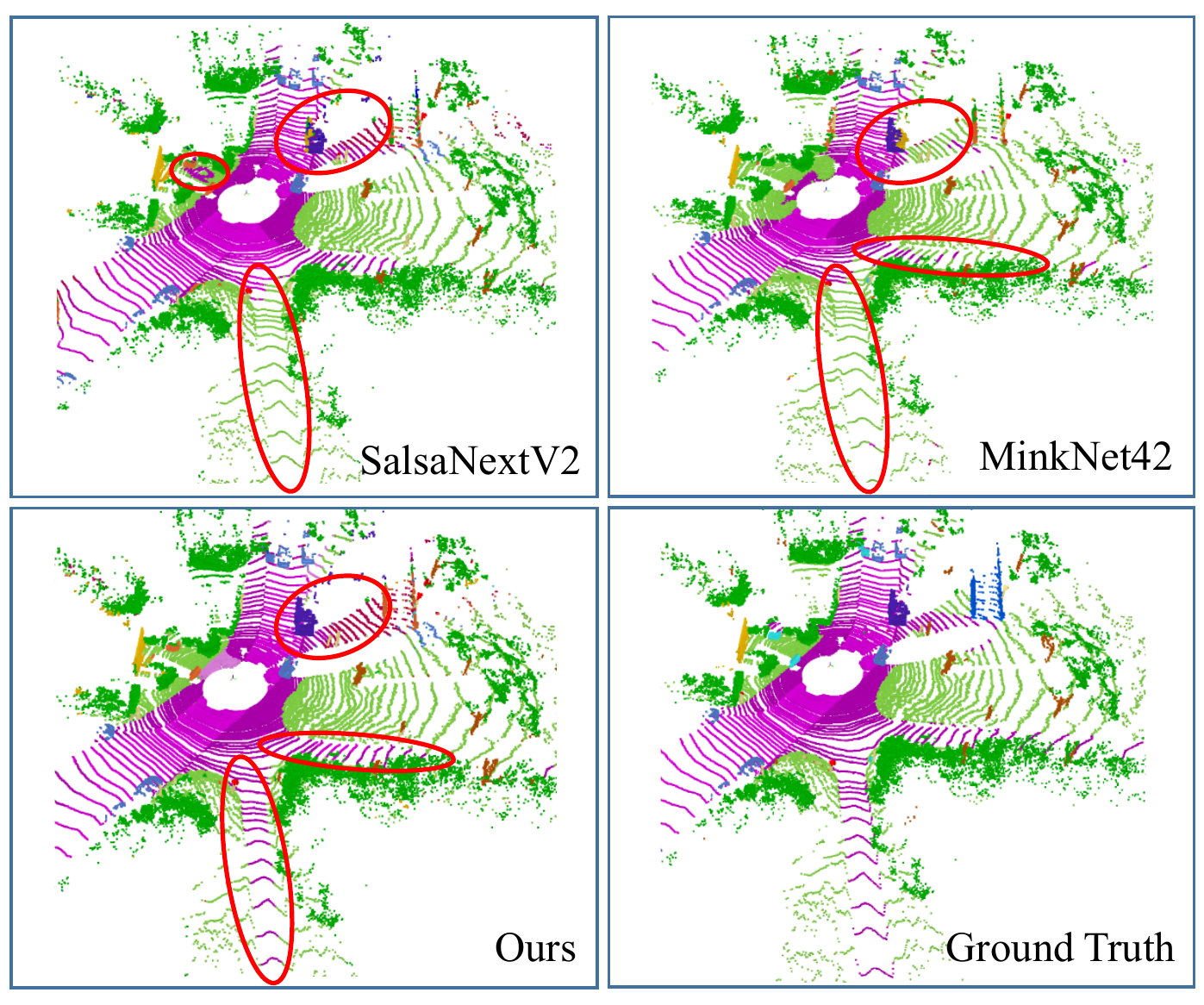}
    \caption{Comparison of our proposed method with SalsanextV2 \cite{cortinhal2020salsanext} and MinkNet42 \cite{choy20194d} on SemanticKITTI benchmark \cite{DBLP:conf/iccv/BehleyGMQBSG19}.}
    \label{fig:cover_image}
    \vspace{-10px}
\end{figure}

Over the past decade, deep learning approaches inspired research in many application such as segmentation due to their success compared to traditional approaches, where handcrafted features are required. While most of these approaches focused on 2D image content \cite{kendall2015bayesian} and reached to a mature stage, fewer contributions have addressed the 3D point cloud semantic segmentation \cite{wu2018squeezeseg}, which is essential for robust scene understanding of autonomous driving.
This is partially due to the fact that camera images provide dense representation in a structured content, whereas LiDAR point clouds are unstructured, sparse and have non-uniform densities across the scene. This characteristics of LiDAR data makes it difficult to extract semantic information in $3$D scenes, although LiDAR point clouds provide more accurate distance measurements of 3D scenes, therefore are beneficial for perception task.
As opposed to traditional approaches, deep learning segmentation networks rely on large quantity of manually annotated dataset which poses additional challenge for research communities. However, due to availability of a large-scale SemanticKITTI dataset \cite{DBLP:conf/iccv/BehleyGMQBSG19}, $3$D LiDAR semantic segmentation task became a hot topic and inspired researchers to propose several methods.

Prior works approach LiDAR segmentation task by processing the input data either in the format of 3D point clouds \cite{Qi_2017_CVPR, qi2017pointnet++} or 2D grids \cite{wu2018squeezeseg, milioto2019rangenet++, aksoy2019salsanet}, e.g., range image,
bird-eye-view. The range-based methods first projects the $3$D point cloud into range-images using spherical projection to benefit from $2$D convolution operation. In particular, this approach faces information loss as \textasciitilde{}30\% of a scene point cloud will be mapped into the overlapping pixels of range-image during the spherical projection. Therefore, after prediction additional post-processing is required to recover some of the points label back into $3$D. The former approach, however processes the point cloud directly in $3$D space which requires heavy computation. 
To partially solve this problem, $3$D sparse convolution can be leveraged which suits varying sparsity of LiDAR point cloud.
Although the sparse convolution reduces the computation complexity without spatial geometrical information loss, however small instances with local details can be lost in the multi-layer propagation. This yields to an unstable or failure to differentiate the fine details.

It is therefore essential to ensure the performance of the safety-critical perception system lies within a certain threshold with reliable confidence score in an attempt to perform semantic segmentation. 
In this paper, we present S3Net, a novel end-to-end sparse 3D convolutional NN for LiDAR point cloud semantic segmentation.
It adopts an encoder-decoder backbone that consists of Sparse Intra-channel Attention Module (SIntraAM), and Sparse Inter-channel Attention Module (SInterAM) to emphasize the fine details both within each feature map and among nearby feature maps.
Moreover, Sparse Residual Tower module is introduced to further process the feature map and extract the global features
We trained our network with Geo-aware anisotropic loss to emphasize the semantic boundaries and penalize the noise within each predicted regions leading to a robust prediction. These improvements are shown in Fig \ref{fig:cover_image}. As illustrated, our method successfully predicted most of the classes such as road and car with better prediction on small objects like moving pedestrian and bicyclists. Our experiments show that the proposed method leads to 12\% improvement compared to its baseline counterpart (MinkNet42 \cite{choy20194d}) on SemanticKITTI  \cite{DBLP:conf/iccv/BehleyGMQBSG19} test set and achieves state-of-the-art performance.

The rest of the paper is organized as follows. In Section \ref{sec:relatedwork}, we introduce a brief review of recent works on LiDAR semantic segmentation task. Then, we present the problem formulation followed by the proposed S3Net model with a detailed description of novel network architecture in Section \ref{sec:method}. A comprehensive experimental results including qualitative, quantitative, and ablation studies on a large-scale public benchmark, SemanticKITTI \cite{DBLP:conf/iccv/BehleyGMQBSG19} is demonstrated in Section \ref{sec:results}. Finally, conclusions and future work are presented in Section \ref{sec:conclusion}.

\begin{figure*}[htbp!]
    \centering
    \includegraphics[trim = 0in 0in -1in 0in, clip, width=6.9in]{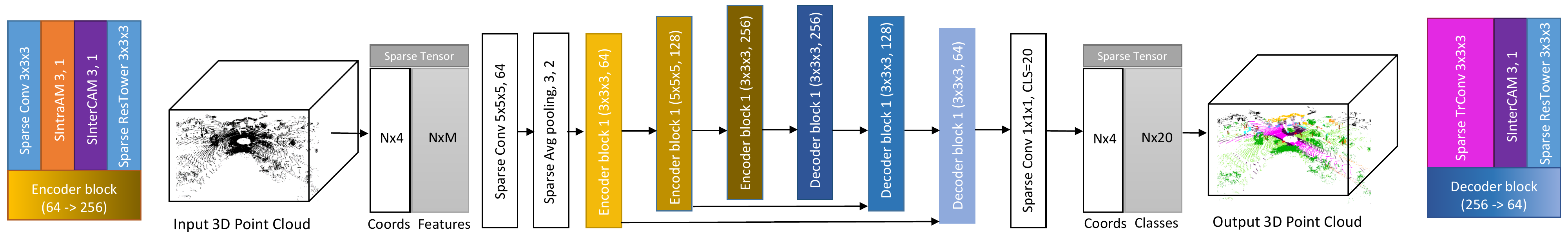}
    \caption[S3Net]{ S3Net Architecture.}
    \label{S3Net}
    \vspace{-10px}
\end{figure*}

\section{RELATED WORK}
\label{sec:relatedwork}
A wide range of deep learning LiDAR semantic segmentation methods have been proposed over the past years. However, based on different processing of input data, they can be broadly classified as projection-based, point-wise, voxel-based, and hybrid methods. Here, we provide a more detailed review of these categories.

\textbf{Projection-based methods} aim to project 3D point clouds into 2D image space either in the top-down Bird-Eye-View \cite{8403277,  Simon2018} or spherical Range-View (RV) format \cite{Milioto2019, cortinhal2020salsanext, xu2020squeezesegv3, 9197193, milioto2019rangenet++, kochanov2020kprnet}. The 2D projection, especially the RV projection, maps the scatter 3D laser points into a 2D image representation which is more dense and compact compared to 3D laser points. Consequently, standard 2D CNN can be leveraged to process these range-images. Therefore, the projection-based methods can achieve real time performance. However, due to the information loss caused by occlusion and density variations, the accuracy of this type of methods is limited.

\textbf{Point-wise methods}, such as PointNet \cite{Qi_2017_CVPR}, PointNet++ \cite{qi2017pointnet++}, KPConv \cite{thomas2019kpconv}, FusionNet \cite{zhang12356deep} and RandLA-Net \cite{Hu_2020_CVPR}, process the raw 3D points without applying any additional transformation or pre-processing. This kind of methods is powerful for small point cloud. While, for the full $360^\circ$ LiDAR scans, the efficiency decreases due to the large memory requirements and slow inference speeds. Usually, the large point cloud needs to be divided into a series of sub-point clouds and feed into the network one by one. The global information of the point cloud is partially lost during this process.

\textbf{Hybrid methods} try to integrate the advantages of the point-wise methods and the projection-based methods. The motivation is to learn the local geometric patterns with point-based method, and the projection-based method is used to gather global context information and accelerate the computation. LatticeNet \cite{rosu2020latticenet} uses a hybrid architecture which leverages the strength of PointNet to obtain low-level features and sparse 3D convolutions to aggregate global context. The raw points are embedded into a sparse lattice, and the convolutions are applied on this sparse lattice. Features on the lattice are projected back onto the point cloud to yield a final segmentation.

Considering the imbalanced spatial distribution of the LiDAR point clouds, PolarNet \cite{zhang2020polarnet} voxelizes raw point cloud in a BEV polar grid, and the points in each grid is produced by a learnable simplified PointNet. 3D-MiniNet \cite{alonso20203d} groups points into view frustums. Point-wise method is implemented in each view frustum, and the global context information is mainly extracted with the MiniNet backbone, which is a 2D CNN operating on range image.

Recently, SPVNAS \cite{tang2020searching} proposed a sparse point-voxel convolution, which includes two branches. The point-based branch is used to capture fine details of the point cloud, and sparse voxel-based branch focuses on global contexts.

\textbf{Voxel-based approaches} transform a point cloud into 3D volumetric grids in which 3D convolutions are used. The early works \cite{7900038} \cite{Gan_2020} directly converted a point cloud into $3$D occupancy grids, then used a standard $3$D CNN for voxel-wise segmentation, and all points within a voxel were assigned the same semantic label as the voxel. However, the accuracy of this method is limited by granularity of the voxels. SEGCloud \cite{Tchapmi2017} generates a coarse voxel prediction by a 3D-FCNN, then maps the result back to point cloud with a deterministic trilinear interpolation. Cylinder3D \cite{zhou2020cylinder3d} transforms point cloud into polar coordinate-system with dimension decomposition to explore high rank context information, VV-Net \cite{meng2019} embeds subvoxels in each voxel and uses a variational autoencoder to capture the point distribution within each voxel. This provides both the benefits of regular structure and capturing the detailed distribution for learning algorithms.

Theoretically, $3$D voxel grid projection is a natural extension of the 2D convolution concept. By using the $3$D voxel grid built in a three-dimensional Euclidean space, the shift invariant (or space invariant) is guaranteed. However, due to high computational complexity and the memory usage, the 3D CNN is far less widely used than the 2D CNN. Fortunately, the computation of high-dimension CNN can be accelerated remarkably by utilizing the space sparsity of the data. B. Graham et al. \cite{8579059} defined a submanifold sparse convolution which can avoid the sparsity reduction caused by standard convolution operation. C. Choy et al. \cite{choy20194d} developed an auto differentiation library for sparse tensors and the generalized sparse convolution, which is called Minkowski engine.
The sparse auto differentiation library reduces the computation complexity while keeping the spatial geometrical information. This benefits the understanding of large outdoor scenes. However, since sparse convolution uses nearest-neighbor search and local down-sampling, small objects with local details can be lost in the multi-layer propagation, leading to an unstable or failure to differentiate the fine details.

\section{METHOD}
\label{sec:method}
In this section, we first introduce the problem formulation followed by a brief review of sparse convolution. Next, the proposed method, S3Net, is presented with detailed description of the network architecture along with network optimization details.

\subsection{Problem formulation}

Given a set of unordered point clouds $P = \{ p_{i} \}$ with $p_{i} \in \mathbb{R}^{d_{in}}$ and $i = 1, ... , N$, where $N$ represents the number of points in an input point cloud scan and $d_{in}$ is the dimensionality of the input features at each point, i.e., coordinates information, colors, etc. Our objective is to predict a set of point-wise label $Y = \{y_i\}$ with $y_{i} \in \mathbb{R}^{d_{out}}$, where $d_{out}$ denotes the class of each input point, and each point $p_{i}$ corresponds one-to-one with $y_{i}$. Thus, S3Net can be modeled as a bijective function $ \mathfrak{F}: P \mapsto Y$, such that the difference between $Y$ and the true labels, $\mathcal{Y}$, are minimized.

\subsection{Sparse Convolutin}
Despite the emerging datasets and technological advancements, processing point cloud still remains challenging due to non-uniform and long-tailed distribution of LiDAR points in a scene. However, one can use sparse convolution which resembles the standard convolution and suits the sparse 3D point cloud.

To this end, we adopt Minkowski Engine \cite{choy20194d}, a sparse tensors auto-differentiation framework, as the foundational building blocks of Sparse convolution of S3Net.

\begin{figure}[htbp!]
    \centering
    \includegraphics[width=0.49\textwidth]{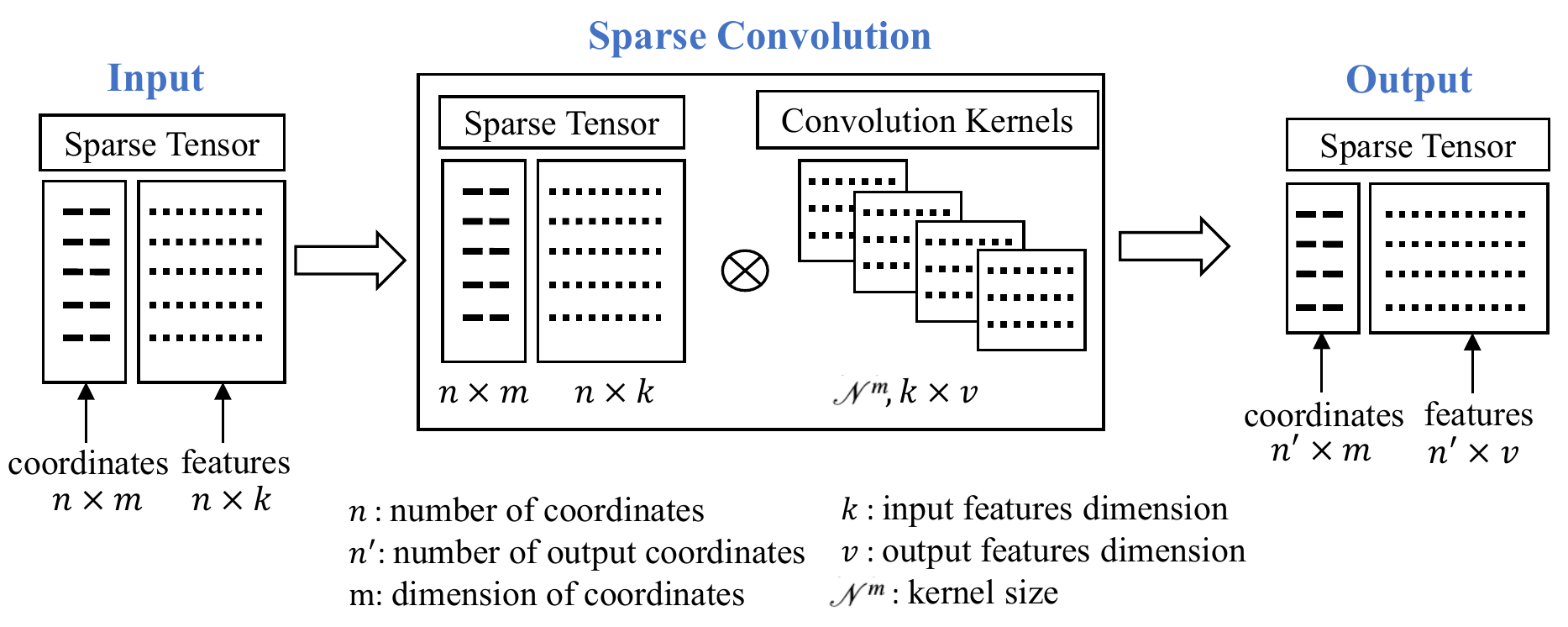}
    \caption{Sparse Convolution Overview}
    \label{fig:Sparse-Convolution-Overview}
\end{figure}

The Sparse convolution is the generalized convolution in arbitrary dimension and arbitrary kernel definitions. In particular, instead of using a dense kernel matrix to define the receptive field, one can define a hashmap to collect adjacent points. This yields to an arbitrary number of coordinates and arbitrary offset of these coordinates within the receptive field.

As shown in Fig. \ref{fig:Sparse-Convolution-Overview}, a sparse tensor $x^{in}=[C_{n \times m}, F_{n \times k}]$ is represented by the input coordinate matrix $C_{n \times m}$ and its corresponding feature matrix $F_{n \times k}$. The generalized sparse convolution can be defined as, 

\begin{align}
x_{u}^{out} = \sum_{i \in \mathcal{N}^m(u, \mathcal{C}^{in})} W_{i} x_{u+i}^{in} \quad \text{for} \quad u \in \mathcal{C}^{out}
\end{align}
where $x_u$ represents the current point that the sparse convolution is applied on, $\mathcal{C}^{in}_{n\times m}$ and $\mathcal{C}^{out}_{n' \times m}$ ($\subset \mathcal{C}^{in}_{n\times m}$) are the input and output coordinate set, respectively. $\mathcal{N}^{m}$ is a set of offsets that define the shape of a kernel.
$W_{i}$ is the $i$th offset weight matrix with shape $(k\times v)$ corresponding to $x_{u+i}^{in}$, where $x_{u+i}^{in}$ is the $i$th adjacent input point ($[C_{u+i}, F_{u+i}]$). Each multiplication projects the feature vector of each coordinate into a new dimensional space according to the weight $W_{i}$. The summation is aggregating the local information into one feature vector which belongs to $x_u^{out}$.

\subsection{3D features}
\label{normal}

Normal features provide additional directional information which help the network differentiate the fine details of objects.
To reduce the heavy computation of 3D point cloud in $3$D space, we first project the point cloud into a 2D range image, in which the width is the azimuth of each scan point and the height is the scan beam number. Then, we derive the 3-channel normal map ($n_x, n_y, n_z$), as expressed by,
\begin{equation}
    n_{x, y, z} = \frac{cat(d_x, d_y)}{\sqrt{d^2_x + d^2_y + 1}}
\end{equation}
where $cat(.)$ denotes the concatenation operation, and $d_{x}$ and $d_{y}$ are the $x$ and $y$ directional gradient of the range image $d$. 
The features of each pixel in the generated 3-dimensional normal surface are mapped back into $3$D points using a mapping table. To recover the points information, mapped on the same range-image pixel, identical features are assigned.

\subsection{Network Architecture}

The block diagram of the proposed S3Net is illustrated in Fig. \ref{S3Net}. The model takes in raw 3D LiDAR point cloud as input and effectively outputs semantic masks to address the road scene understanding problem that is an essential prerequisite for autonomous vehicles. We first generate 3D normal features for every single LiDAR scan as described in section \ref{normal}. Then, for a raw point cloud, we construct sparse tensor to be processed by the network. The sparse tensor representation of data is obtained by converting the points into coordinates and features. The network consists of an encoder-decoder structure with series of novel SIntraAM, SInterAm, and Sparse ResTower modules which process the point cloud using sparse convolution, as shown in Fig.\ref{S3Net}. 
Below, we present the detailed explanation of our method. 

\subsection{Intra-channel}

\begin{figure}[htbp!]
    \centering
    \includegraphics[width=0.45\textwidth]{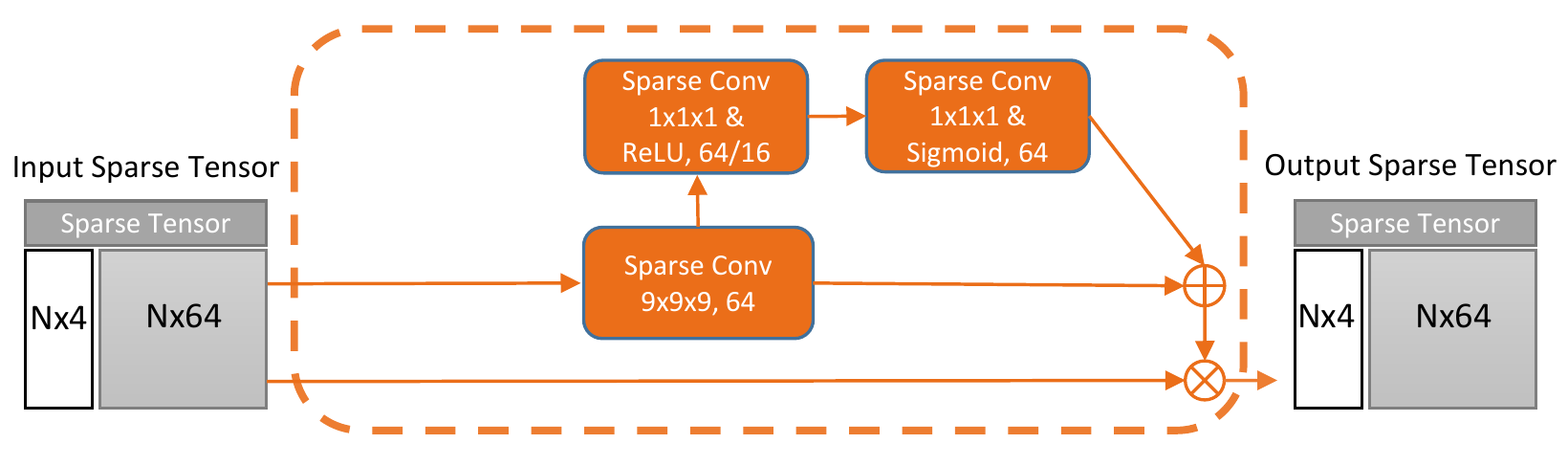}
    \caption{SIntraAM}
    \label{fig:SIntraAM}
    \vspace{-10px}
\end{figure}

To overcome the local information loss as a result of using sparse convolution, we propose attention module, inspired by CAM \cite{yu2015multi} to emphasize critical features within each feature map. 
As illustrated in Fig. \ref{fig:SIntraAM}, we design a new Sparse Intra-channel attention module (SIntraAM) in our method in 3D domain to apply attention over the context of the feature maps. CAM module is implemented by aggregating the contextual information using max pooling. However, it introduces loss of information. To overcome this information loss, we used sparse $3$D convolution to learn better feature representation for attention mask. A stabilizing step is added for the final generated mask in 3D by applying element-wise multiplication of the attention mask with the input sparse tensor.

This block can algebraically be described by the back-propagation equation:

\begin{equation}
    x^{k} = x^{k+1} + x^{k+1}(\frac{\partial \mathcal{F}^{k}}{\partial x} + x^{k+2}\frac{\partial \mathcal{F}^{k+1}}{\partial x})
\end{equation}
where $x^{k}$ and $\mathcal{F}$ represent layer $k_{th}$ output and its corresponding convolution layer. These residual connections ensure that when $\partial \mathcal{F}/ \partial x = 0$, the error signals are still able to pass through (unmodified) successor layers (e.g. $x^{k+1}$ or $x^{k+2}$). As a result, the jacobian matrix is close to identity which increases the network stability.

Encoder convolutions extract different features from the input data. Decoder convolutions further processes the features generated from the encoder and up-samples those local and global feature maps to predict the final labels for full point cloud. The features generated by decoder are filled with artifacts (at least at early stages of training). Preliminary analysis of the system determined the design choice of the decoder blocks not to include the Sparse Intra-channel attention module.

\subsection{Inter-channel}

\begin{figure}[htbp!]
    \centering
    \includegraphics[width=0.45\textwidth]{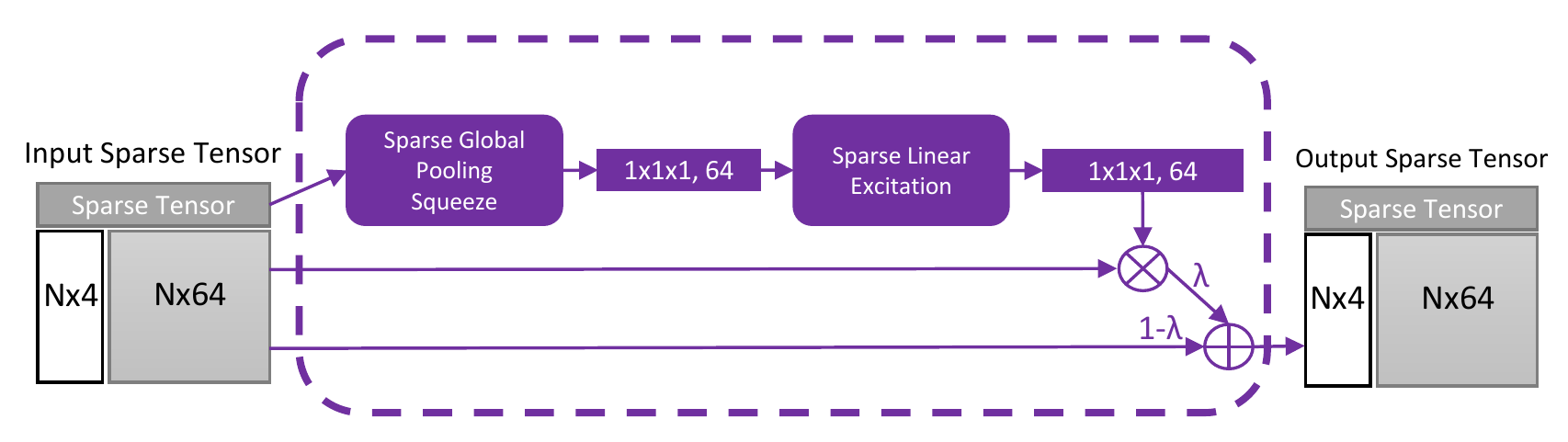}
    \caption{SInterAM}
    \label{fig:SInterAM}
    \vspace{-10px}
\end{figure}

To emphasize the channel-wise feature map in each layer, we designed a new Sparse Inter-channel attention module (SInterAM) in our method in 3D domain, inspired by Squeeze Re-weight module \cite{hu2018squeeze} from image domain. The block diagram of SInterAM is shown in Fig. \ref{fig:SInterAM}. More specifically, a sparse global average pooling squeeze layer is applied on the input sparse tensor to obtain the global information. The feature map is then passed to a sparse linear excitation block to generate channel-wise dependencies. Next, we apply scaling factor to the output feature map of sparse linear excitation. In the final step, we introduce a damping factor $\lambda$ after the element-wise multiplication to regularize the final scaled feature map. The value of $\lambda$ is empirically set to 0.35.

\begin{figure}[htbp!]
    \centering
    \includegraphics[width=0.45\textwidth]{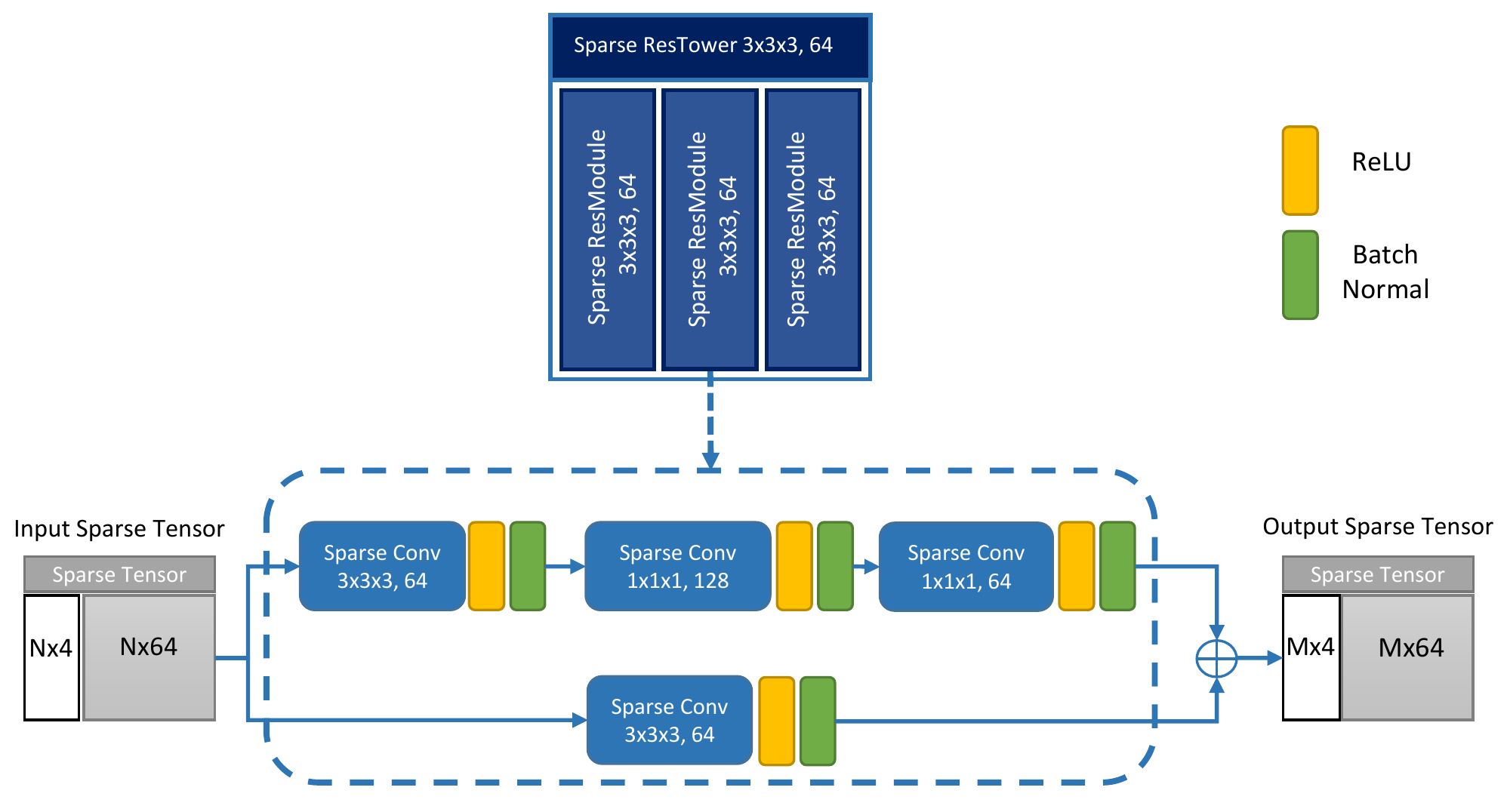}
    \caption{SparseResTower}
    \label{fig:SparseResTower}
    \vspace{-10px}
\end{figure}

Sparse Residual Tower is introduced to further process the feature maps and extract the abstract/global features. It consists of series of sparse convolution block followed by ReLU activation and Batch normalization layer. As illustrated in Fig. \ref{fig:SparseResTower}, the coordinates of the input and output sparse tensor of the Sparse ResModules are mismatched. Thus, we designed a skip connection using a sparse convolution to align these coordinates and add the output features per point. The ResTower block of the encoder and decoder are built upon using 3 and 2 of the ResModules, respectively.

\subsection{loss function}

We introduced the geo-aware anisotropic loss \cite{li2019depth} combined with cross-entropy loss  \cite{zhang2018generalized, panchapagesan2016multi} to determine the importance of different positions within the scene. It is beneficial for classification application such as semantic segmentation to recover the fine details like the boundaries of objects and the corners of the scene.

The weighted cross entropy loss, ${L}_{wce}(y,\hat{y})$, can be calculated as,

\begin{align}
{L}_{wce}(y,\hat{y})= - \sum_{c \in C} \alpha_{c} \> y_{c} \> \log P(\hat{y}_{c}),\quad  \alpha_{c}= 1/\sqrt{(f_{c})}
\end{align}
where $y_{c}$ and $P(\hat{y}_{i})$ are the ground truth and the model output probability for each class $c \in C$, respectively. $\alpha_{c}$ denotes the frequency of each class and $f_{c}$ is the frequency of the class labels in the whole dataset.

Th second term is geo-aware anisotropic loss and can be computed by,

\begin{align}
    {L}_{geo}(y,\hat{y}) = -\frac{1}{N}\sum_{i,j,k}\sum_{c=1}^{C}\frac{M_{LGA}}{\Phi}y_{ijk,c}log\hat{y}_{ijk, c}
\end{align}
where $N$ is the neighborhood of the current voxel cell located at $i, j, k$. $c \in C$ is the current semantic classes. $y$ is the ground truth label and $\hat{y}$ is the predicted label. This equation is a regularized version of cross entropy where the regualizer is $M_{LGA} = \sum^{\Phi}_{\phi=1}(c_p \oplus c_{q_{\phi}})$, defined by Li \emph{et al} \cite{li2019depth}. We normalized local geometric anisotropy within the sliding window $\Phi$ of the current voxel cell $p$. $q_{\phi}$ is one of the voxel cell located in the sliding window $\Phi$, where $\phi \in {1,2,3,...,\Phi}$, and $\oplus$ is the exclusive operation. $L_{geo}$ smoothens up the loss manifold for the object boundaries and noisy prediction of multiple semantic classes at intersection corners and penalizes the prediction errors for those local details.

Thus, the total loss used to train the proposed network is calculated as,
\begin{align}
    {L}_{tot}(y,\hat{y}) = \lambda_{1} {L}_{wce}(y,\hat{y}) + \lambda_{2} {L}_{geo}(y,\hat{y}) 
\end{align}
where $\lambda_{1}$ and $\lambda_{2}$ represent the weights of weighted cross entropy and geo-aware anisotropic loss, respectively. In our experiments, we found that $\lambda_{1} = 0.75$ with $\lambda_{2} = 0.25$ achieves best performance.

\begin{table*}[htb]
{\Huge
\centering
\resizebox{1.75\columnwidth}{!}{
\begin{tabular}{l|cccccccccccccccccccc}
\hline 
Method & \begin{sideways} Mean IoU \end{sideways} 
& \begin{sideways} Car \end{sideways} 
& \begin{sideways} Bicycle \end{sideways} 
& \begin{sideways} Motorcycle \end{sideways} 
& \begin{sideways} Truck \end{sideways} 
& \begin{sideways} Other-vehicle \end{sideways} 
& \begin{sideways} Person \end{sideways} 
& \begin{sideways} Bicyclist \end{sideways} 
& \begin{sideways} Motorcyclist \end{sideways} 
& \begin{sideways} Road \end{sideways} 
& \begin{sideways} Parking \end{sideways} 
& \begin{sideways} Sidewalk \end{sideways} 
& \begin{sideways} Other-ground \end{sideways} 
& \begin{sideways} Building \end{sideways} 
& \begin{sideways} Fence \end{sideways} 
& \begin{sideways} Vegetation \end{sideways} 
& \begin{sideways} Trunk \end{sideways} 
& \begin{sideways} Terrain \end{sideways} 
& \begin{sideways} Pole \end{sideways} 
& \begin{sideways} Traffic-sign \end{sideways}\\ 
\hline

S-BKI \cite{Gan_2020} 
& $51.3$ & $83.8$ & $30.6$ & $43.0$ & $26.0$ & $19.6$ & $8.5$ & $3.4$ & $0.0$ & $92.6$ & $65.3$ & $77.4$ & $30.1$ & $89.7$ & $63.7$ & $83.4$ & $64.3$ & $67.4$ & $58.6$ & $67.1$  \\
RangeNet \cite{milioto2019rangenet++} 
& $52.2$ & $91.4$ & $25.7$ & $34.4$ & $25.7$ & $23.0$ & $38.3$ & $38.8$ & $4.8$ & $91.8$ & $65.0$ & $75.2$ & $27.8$ & $87.4$ & $58.6$ & $80.5$ & $55.1$ & $64.6$ & $47.9$ & $55.9$  \\
LatticeNet  \cite{rosu2019latticenet} 
& $52.9$ & $92.9$ & $16.6$ & $22.2$ & $26.6$ & $21.4$ & $35.6$ & $43.0$ & $46.0$ & $90.0$ & $59.4$ & $74.1$ & $22.0$ & $88.2$ & $58.8$ & $81.7$ & $63.6$ & $63.1$ & $51.9$ & $48.4$ \\
RandLa-Net \cite{hu2019randla} 
& $53.9$ & $94.2$ & $26.0$ & $25.8$ & $40.1$ & $38.9$ & $49.2$ & $48.2$ & $7.2$ & $90.7$ & $60.3$ & $73.7$ & $20.4$ & $86.9$ & $56.3$ & $81.4$ & $61.3$ & $66.8$ & $49.2$ & $47.7$  \\
PolarNet \cite{zhang2020polarnet} 
& $54.3$ & $93.8$ & $40.3$ & $30.1$ & $22.9$ & $28.5$ & $43.2$ & $40.2$ & $5.6$ & $90.8$ & $61.7$ & $74.4$ & $21.7$ & $90.0$ & $61.3$ & $84.0$ & $65.5$ & $67.8$ & $51.8$ & $57.5$  \\
MinkNet42 \cite{choy20194d} 
& $54.3$ &  $94.3$ & $23.1$ & $26.2$ & $26.1$ & $36.7$ & $43.1$ & $36.4$ & $7.9$ & $91.1$ & $63.8$ & $69.7$ & $29.3$ & $92.7$ & $57.1$ & $83.7$ & $68.4$ & $64.7$ & $57.3$ & $60.1$ \\ 
3D-MiniNet \cite{alonso20203d} 
& $55.8$ & $90.5$ & $42.3$ & $42.1$ & $28.5$ & $29.4$ & $47.8$ & $44.1$ & $14.5$ & $91.6$ & $64.2$ & $74.5$ & $25.4$ & $89.4$ & $60.8$ & $82.8$ & $60.8$ & $66.7$ & $48.0$ & $56.6$ \\
SqueezeSegV3 \cite{xu2020squeezesegv3} 
& $55.9$ & $92.5$ & $38.7$ & $36.5$ & $29.6$ & $33.0$ & $45.6$ & $46.2$ & $20.1$ & $91.7$ & $63.4$ & $74.8$ & $26.4$ & $89.0$ & $59.4$ & $82.0$ & $58.7$ & $65.4$ & $49.6$ & $58.9$ \\
Kpconv \cite{thomas2019kpconv} 
& $58.8$ & $96.0$ & $30.2$ & $42.5$ & $33.4$ & $44.3$ & $61.5$ & $61.6$ & $11.8$ & $88.8$ & $61.3$ & $72.7$ & $31.6$ & $90.5$ & $64.2$ & $84.8$ & $69.2$ & $69.1$ & $56.4$ & $47.4$ \\ 
SalsaNext \cite{cortinhal2020salsanext}
& $59.5$ & $91.9$ & $48.3$ & $38.6$ & $38.9$ & $31.9$ & $60.2$ & $59.0$ & $19.4$ & $91.7$ & $63.7$ & $75.8$ & $29.1$ & $90.2$ & $64.2$ & $81.8$ & $63.6$ & $66.5$ & $54.3$ & $62.1$    \\
FusionNet   \cite{zhang12356deep}
& $61.3$  & $95.3$ & $47.5$ & $37.7$ & $\textbf{41.8}$ & $34.5$ & $59.5$ & $56.8$ & $11.9$ & $91.8$ & $68.8$ & $77.1$ & $30.8$ & $\textbf{92.5}$ & $\textbf{69.4}$ & $84.5$ & $69.8$ & $68.5$ & $60.4$ & $66.5$\\
Cylinder3D  \cite{zhou2020cylinder3d}
& $61.8$ & $\textbf{96.1}$ & $54.2$ & $47.6$ & $38.6$ & $\textbf{45.0}$ & $65.1$ & $63.5$ & $13.6$ & $91.2$ & $62.2$ & $75.2$ & $18.7$ & $89.6$ & $61.6$ & $85.4$ & $69.7$ & $69.3$ & $\textbf{62.6}$ & $64.7$\\
KPRNet    \cite{kochanov2020kprnet}
& $63.1$ & $95.5$ & $54.1$ & $47.9$ & $23.6$ & $42.6$ & $65.9$ & $65.0$ & $16.5$ & $\textbf{93.2}$ & $\textbf{73.9}$ & $\textbf{80.6}$ & $30.2$ & $91.7$ & $68.4$ & $\textbf{85.7}$ & $\textbf{69.8}$ & $\textbf{71.2}$ & $58.7$ & $64.1$\\
 
\hline
 \textbf{S3Net} [Ours] 
& $\textbf{66.8}$ & $93.7$ & $53.5$ & $\textbf{80.0}$ & $32.0$ & $34.9$ & $\textbf{74.0}$ & $\textbf{80.7}$ & $\textbf{80.7}$ & $89.3$ & $57.1$ & $71.8$ & $\textbf{36.6}$ & $91.5$ & $64.2$ & $76.0$ & $65.5$ & $59.8$ & $58.0$ & $\textbf{70.0}$  \rule{0pt}{2.3ex}\\

\hline
\end{tabular}
}
\caption[S3Net]{IoU results on the SemanticKITTI test dataset. FPS measurements were taken using a single GTX 2080Ti GPU, or approximated if a runtime comparison was made on another GPU.}
\label{bigtable}}
\end{table*} 

\section{EXPERIMENTS}
\label{sec:results}

\subsection{Setting}
\subsubsection{Dataset}

We evaluated our model on a large-scale LiDAR semantic segmentation dataset for autonomous vehicle, SemanticKITTI \cite{DBLP:conf/iccv/BehleyGMQBSG19}. This dataset is based on KITTI Odometry benchmark which contains $22$ driving sequences, where the first $11$ sequences are labelled with each point belonging to one of $22$ distinct classes and available for public. However, the labels for other $10$ sequences are reserved for the competition and are not released. More specifically, it comprised of 23,201 LiDAR scans for training and 20,351 for test set. As a common practice, sequence $08$ is suggested to be considered as a validation set at the training time and we followed this split in our training.
The evaluation of the methods are based on Jaccard Index, also known as mean intersection-over-union (IoU) \cite{milioto2019rangenet++} metric.

\subsubsection{Experiment Settings} 

We trained our model with Adam optimizer with learning rate of 0.001, momentum of 0.9 and weight decay of 0.0005 for 120 epochs. We used a exponential scheduler \cite{li2019exponential} with a decay rate of 0.9 and the update period of 10 epochs. We trained our model using 8 V100 GPU in parallel for around 200 hours.

\subsection{Quantitative Evaluation}
In this section, we provide the results of our proposed method along with all the existing semantic segmentation methods with mIoU of higher than 50\% on SemanticKITTI benchmark. As illustrated in Table \ref{bigtable}, S3Net achieves state-of-the-art performance on SemanticKITTI test set benchmark, especially for the small objects like motorcycle and motorcyclist. At the time of writing this paper, our proposed S3Net outperforms all existing published works such as, SPVNAS \cite{tang2020searching} (66.4\%), KPRNet \cite{kochanov2020kprnet} (63.1\%), Cylinder$3$D \cite{zhou2020cylinder3d} (61.8\%) by 0.4\%, 3.7\%, 5\% in terms of mean IoU score, respectively.
Moreover, our method achieves an improvement of 12.5$\%$ mIoU as opposed to its baseline method \cite{choy20194d}.

\subsection{Qualitative Evaluation}

\begin{figure}[htb]
    \centering
    \includegraphics[width=0.48\textwidth]{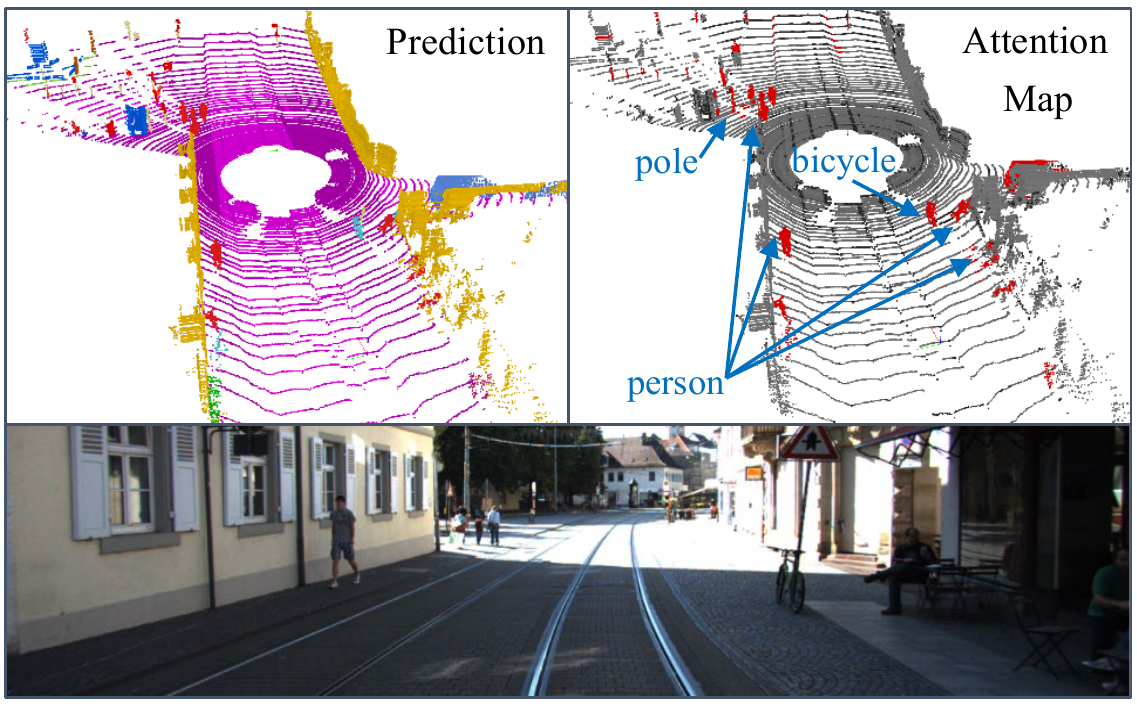}
    \caption{Prediction (top-left), attention map (top-right), and reference image (bottom) on SemanticKITTI test set. S$3$Net attention modules learn to emphasize the fine details on smaller objects (i.e., person, bicycle, pole, traffic-sign).}
    \label{fig:attention-map}
    \vspace{-10px}
\end{figure}

In Fig. \ref{fig:attention-map} we demonstrate the activated feature map. We normalized feature maps for all 3D point clouds after the last decoder layer. Then, we selected the first 2\% of the points with highest feature values. It can be observed that the network learns the local attention to the small objects like person, pole, bicycles and other small structures like traffic signs.

\begin{figure}[htb]
    \centering
    \includegraphics[width=0.48\textwidth]{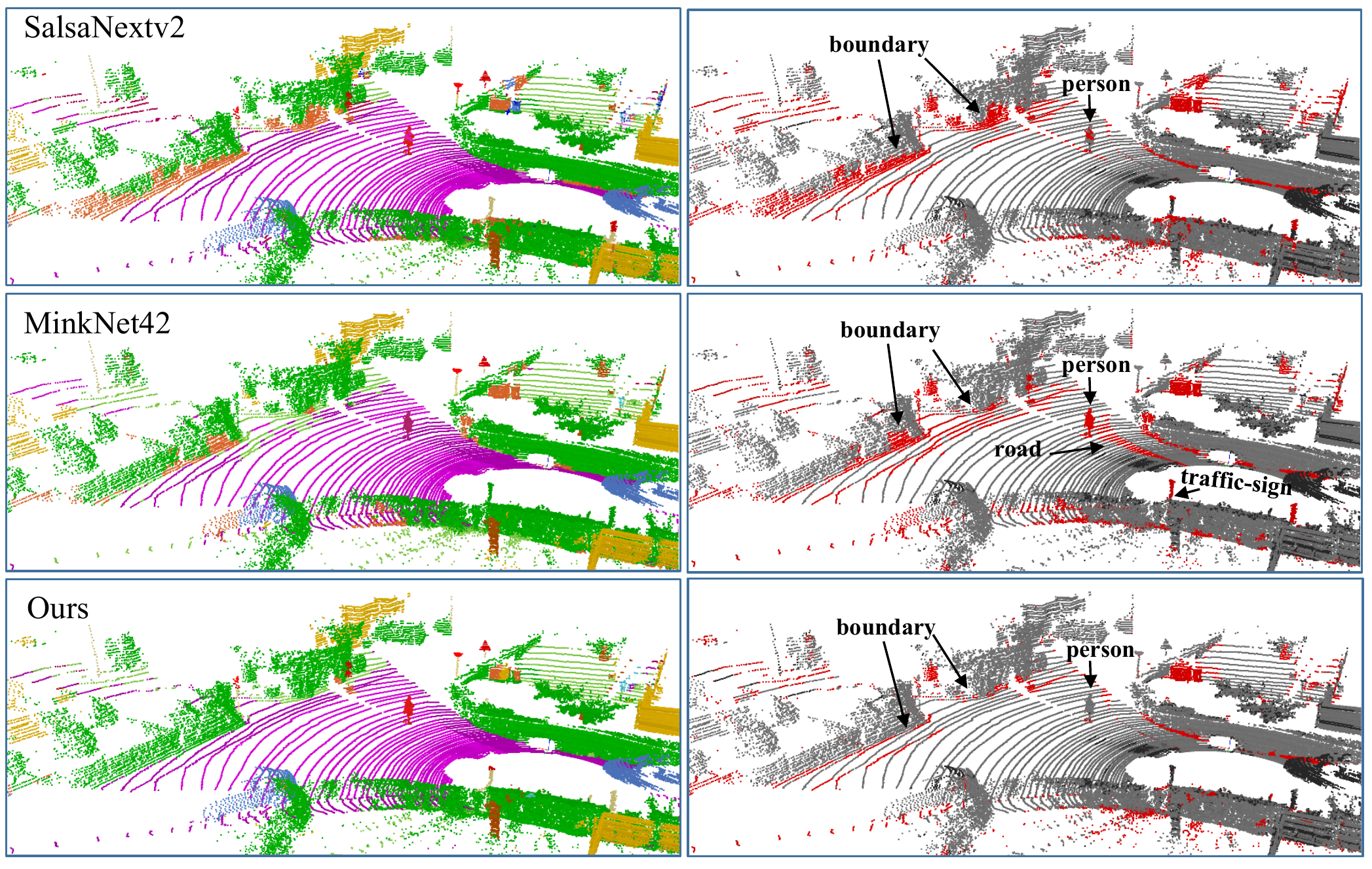}
    \caption{Prediction (left) and error map (right) on SemanticKITTI validation set. SalsanextV2 suffers from bleeding effect due to spherical projection and  MinkNet42 failed to predict person and boundary surface, while ours shows successful semantic predictions.}
    \label{fig:Qualitative_err}
    \vspace{-10px}
\end{figure}

As shown in Fig. \ref{fig:Qualitative_err}, it can be observed that our method has smaller error over the semantic boundaries as well as the small objects like pedestrians and pole-like object.

\subsection{Ablation study}
We provide extensive experiments investigating the contribution of the components proposed in our method, such as leveraging normal features, SInterAM, SIntraAM, SResTower, and geo-aware anisotropic loss (Geo-loss), shown  in Table \ref{tab:Ablation}. 
The experiments are conducted on SemanticKITTI validation set (Sequence 08). It can be observed that introducing SInterAM and SIntraAM modules leads to an increase in accuracy by 2\%, which demonstrates the effectiveness of the attention to the local details. As illustrated, the significant jump in accuracy is due to introducing the Sparse Res Tower by additional 3.3\% improvement, due to residual layers promoting the gradient signal to previous layers to update feature extractors.
It is worth noting that the geo-loss further boosts the performance by applying attention to the boundaries information of each semantic classes. Overall, our model significantly improves the accuracy compared to the baseline by a large margin (8.5\%).

\begin{table}[h]
\begin{center}
\scalebox{0.8}{
\begin{tabular}{ c|cccccc c  }
\hline 
\multicolumn{1}{c|}{\textbf{Architecture}} &
\multicolumn{1}{c}{\textbf{\begin{tabular}[c]{@{}c@{}}Norm\\ Feature\end{tabular}}}  &  
\multicolumn{1}{c}{\textbf{SInterAM}} & 
\multicolumn{1}{c}{\textbf{SIntraAM}} &
\multicolumn{1}{c}{\textbf{SResTower}} &
\multicolumn{1}{c|}{\textbf{Geo-loss}} &
\multicolumn{1}{ c}{\textbf{mIoU}} \\

 \hline \hline 
\multirow{1}{*}{Baseline}
&     &   &   &    &    \multicolumn{1}{c|}{ } & 59.8 \rule{0pt}{3ex}\\\hline

\multirow{5}{*}{S3Net}
& \checkmark &   &   &   &    \multicolumn{1}{c|}{ }  & 62.2  \rule{0pt}{3ex}\\ 

& \checkmark & \checkmark &   &   &   \multicolumn{1}{c|}{ }  & 62.1  \rule{0pt}{3ex}\\ 

& \checkmark & \checkmark &\checkmark  &   &    \multicolumn{1}{c|}{ }  & 64.1  \rule{0pt}{3ex}\\ 

& \checkmark & \checkmark &\checkmark  & \checkmark  &   \multicolumn{1}{c|}{ }  & 67.4  \rule{0pt}{3ex}\\

& \checkmark & \checkmark &\checkmark  & \checkmark  & \multicolumn{1}{c|}{\checkmark}  & 68.3  \rule{0pt}{3ex}\\ \hline

\end{tabular}}
\end{center}
\caption{Ablation study of the proposed method vs baseline evaluated on SemanticKITTI dataset validation (seq 08). }
\label{tab:Ablation}
\end{table}

\section{CONCLUSIONS}
\label{sec:conclusion}

In this work, we have presented an end-to-end sparse $3$D CNN model for  LiDAR  point  cloud  semantic  segmentation. The proposed method, S$3$Net, has an encoder-decoder architecture which is built on the basis of sparse convolution. We introduced novel Sparse Intra-channel Attention Module (SIntraAM), and Sparse Inter-
channel  Attention  Module  (SInterAM)  to  capture  the  critical features  in  each  feature  map  and  channels  and  to emphasize them  for  better  representation. To optimize the network, we leveraged the geo-aware  anisotropic loss, in addition to weighted cross entropy, to  emphasize  the  semantic boundaries which lead to improved accuracy.
We have demonstrated a thorough qualitative and quantitative evaluation on a large-scale public benchmark, SemanticKITTI. 
Our experimental results show that the proposed method leads to a large improvement (12\%) as opposed to its baseline counterpart (MinkNet42 \cite{choy20194d}) on SemanticKITTI test set and achieves state-of-the-art performance.
Our method lays ground for further development of such promising designs to solve not only the problem of LiDAR semantic segmentation but also other downstream perception tasks, such as object detection and tracking.

\bibliographystyle{IEEEtran}
\bibliography{summarybib}

\end{document}